\documentclass[letterpaper, 10 pt, conference]{ieeeconf}  %

\IEEEoverridecommandlockouts                              %

\usepackage{graphics} %
\usepackage{multirow}
\usepackage{booktabs}
\usepackage{hyperref}
\usepackage{subcaption}
\usepackage{epsfig} %
\usepackage{comment}
\usepackage{mathptmx} %
\usepackage{amsmath} %
\usepackage{amssymb}  %
\usepackage{color}
\usepackage{paralist}
\usepackage[normalem]{ulem} %

\DeclareMathOperator{\dgo}{DemoGrasp}
\newcommand{\demograsp}[0]{$\dgo$}

\DeclareMathOperator{\6gn}{6DoF\,GraspNet}
\newcommand{\graspn}[0]{$\6gn$}

\renewcommand{\vec}[1]{\mathbf{#1}}

\definecolor{sven}{rgb}{0,1,1}

\definecolor{ben}{rgb}{0.9,0.,0.5}

\title{\LARGE \bf
DemoGrasp: Few-Shot Learning for Robotic Grasping\\with Human Demonstration
}

\author{Pengyuan Wang$^{1,*}$, Fabian Manhardt$^{1,*}$, Luca Minciullo$^{2}$, Lorenzo Garattoni$^{2}$, \\ Sven Meier$^{2}$, Nassir Navab$^{1}$ and Benjamin Busam$^{1}$%
\thanks{$^{1}$ Technical University of Munich {\tt\small \{first.last\}@tum.de}}
\thanks{$^{2}$ Toyota Motor Europe {\tt\small \{first.last\}@toyota-europe.com}}
\thanks{$^{*}$ Equal Contribution}}

\begin{document}

\maketitle

\thispagestyle{empty}
\pagestyle{empty}

\begin{abstract}
The ability to successfully grasp objects is crucial in robotics, as it enables several interactive downstream applications.
To this end, most approaches either compute the full 6D pose for the object of interest or learn to predict a set of grasping points.
While the former approaches do not scale well to multiple object instances or classes yet, the latter require large annotated datasets and are hampered by their poor generalization capabilities to new geometries.
To overcome these shortcomings, we propose to teach a robot how to grasp an object with a simple and short human demonstration.
Hence, our approach neither requires many annotated images nor is it restricted to a specific geometry.
We first present a small sequence of RGB-D images displaying a human-object interaction.
This sequence is then leveraged to build associated hand and object meshes that represent the depicted interaction.
Subsequently, we complete missing parts of the reconstructed object shape and estimate the relative transformation between the reconstruction and the visible object in the scene.
Finally, we transfer the \emph{a-priori} knowledge from the relative pose between object and human hand with the estimate of the current object pose in the scene into necessary grasping instructions for the robot.
Exhaustive evaluations with Toyota's Human Support Robot (HSR) in real and synthetic environments demonstrate the applicability of our proposed methodology and its advantage in comparison to previous approaches.
\end{abstract}
\section{INTRODUCTION}
Grasping of objects is a fundamental problem in robotics as it enables a numerous applications~\cite{alonso2018current}.
Robotic manipulators are already an integral part in modern workplaces where they are often used for repetitive tasks~\cite{LIANG2020103370}.  While human-robot collaboration can even help in medical applications~\cite{esposito2015cooperative,busam2015stereo}, it is often restricted to cases with clearly defined a priori motion patterns.
When the interaction is more intricate and active manipulation is required, a priori definition becomes non-trivial such as in a less structured domestic environments where service robots typically operate.
Current approaches for robotic grasping lack generalization capabilities as they either concentrate on estimating the object pose~\cite{kleeberger2020survey, sahin2020review} or learn grasping points~\cite{mousavian20196} which require detailed prior information of the object or a large set of annotations.
Just like human hands, robotic grippers and arms have natural limits in their range of motion and have limited degrees-of-freedom, which restrict their possible grasping poses. While the motion models for robotics gripper and human hands can differ greatly, it should be possible to distill information from human manipulation and deduce adequate gripping commands for the target robot from it.
Doing so with a limited set of human demonstrations would enable a robot to imitate the human behaviour and thus enable to seamlessly grasp objects.
\begin{figure}[t!]
    \centering
    \includegraphics[width=\linewidth]{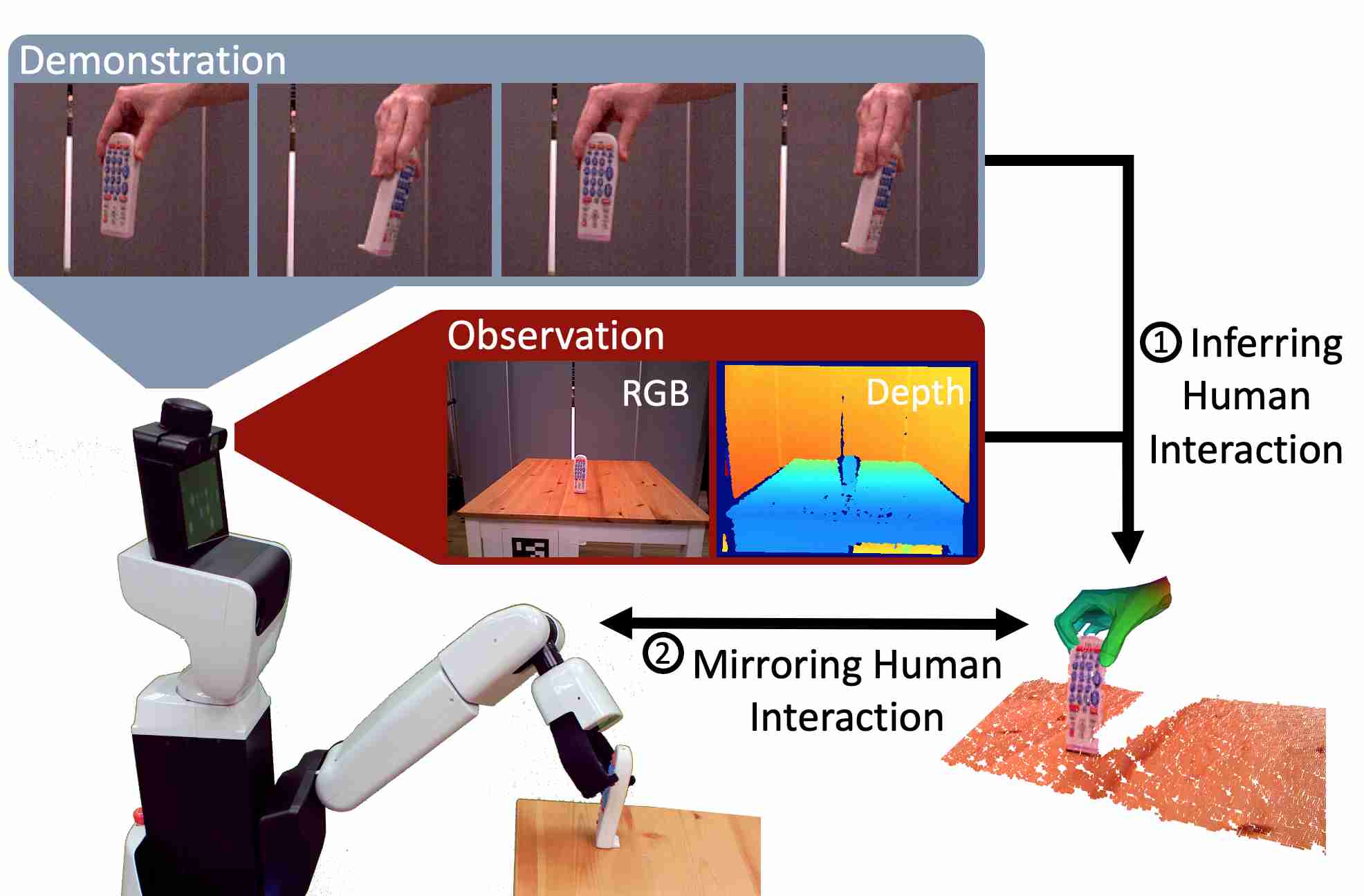}
    \caption{\textbf{DemoGrasp pipeline.} After demonstrating human interaction with the object of interest, our method leverages this knowledge to infer correct human-object interaction for the current observation (1). Subsequently, grasping instructions are derived from the inferred interaction (2).}
    \label{fig:teaser}
\end{figure}

We focus on this imitation where the robot mirrors the human interaction as illustrated in Fig.~\ref{fig:teaser}.
The task can be partitioned into a visual perception and an interpretation part where a human instructor demonstrates the manipulation a priori (Demo) from which the robot deduces the grasping information necessary to manipulate the current scene (Grasp).
Given an adequate mapping from the human hand to the robotic gripper, decomposing the task into these two stages allows our method to scale to a large amount of different grippers.
Eventually, this paves the way to teach the robot by natural human demonstration, which allows to realize a higher level of automation, especially in less structured environments.

During demonstration of the object to the robot (Demo) from various different viewpoints, our method constantly tracks both the hand and the object which are fused into a Truncated Signed Distance Field (TSDF) for 3D reconstruction~\cite{izadi2011kinectfusion, lorensen1987marching}.
Using semantic segmentation of hand and object~\cite{he2017mask}, the reconstruction can be separated and further processed to retrieve a full 3D representation of both object and hand. We then extract the associated 3D hand mesh leveraging the MANO~\cite{romero2017embodied} hand model, which we tightly align with the reconstructed object.
During inference, we then use PPF-FoldNet~\cite{deng2018ppf} to predict if an object is present together with its relative transformation from object to camera space.
The final grasp instructions are then derived from the estimated hand mesh applying the estimated pose.

To summarize, we propose \demograsp, the first learning by demonstration pipeline that can infer robotic grasping pose directly from a short human hand demonstration sequence of RGB-D images. Further, we set up a new synthetic evaluation benchmark based on the HSR to systematically evaluate grasping via demonstration. We make the dataset publicly available to encourage future research in this field.

\section{RELATED WORK}
Methods for robotic grasping can be divided into model-based and model-free approaches~\cite{kleeberger2020survey}. While model-based approaches require a 3D CAD for each object of interest, model-free methods instead directly infer final grasp instructions without any knowledge about the object's geometry.

\subsection{Model-based Grasping}
The task of model-based grasping commonly involves solving for the 6D pose, \emph{i.e.}~3D rotation and 3D translation, of the object of interest~\cite{kehl2017ssd,deng2020self}. Most traditional methods for object pose estimation rely on local image features such as SIFT~\cite{lowe1999object,Romea-2011-7355} or template matching \cite{hinterstoisser2012gradient}. With the rise of consumer depth cameras, the trend shifted towards 6D pose estimation from RGB-D data. Also with the additional depth map, template matching~\cite{hinterstoisser2012} has been used while others proposed to use handcrafted 3D descriptors such as SHOT~\cite{tombari2010unique} or point pair features~\cite{drost2010model,hinterstoisser2016going}, or learn the pose task~\cite{Brachmann2014Learning6O, krull2015learning}.
With the advent of deep learning, 6D pose estimation received another boost in attention as consistently faster and more accurate approaches have been introduced~\cite{hodan2018bop}. In essence, there are three different lines of work for estimating the 6D pose. The first is grounded on 3D descriptors~\cite{qi2017pointnet,deng2018ppf}. These descriptors can be computed for example via metric learning~\cite{wohlhart2015learning} or by means of auto-encoding~\cite{sundermeyer2018implicit}. Other methods directly infer the 6D pose~\cite{xiang2017posecnn,kehl2017ssd}. While \cite{wang2019densefusion,xiang2017posecnn} regress the output pose, Kehl~\emph{et al.}~turn it into a classification problem~\cite{kehl2017ssd}. A few methods also solve for the pose by means of pose refinement~\cite{li2018deepim, manhardt2018deep, labbe2020cosypose}. The last and most prominent branch of works establishes 2D-3D correspondences and optimize for pose using a variant of the P$n$P/RANSAC paradigm~\cite{rad2017bb8, tekin2018real, peng2019pvnet, hodan2020epos}. 

While the accuracy of estimating the 6D pose keeps steadily increasing, it also comes with a heavy burden in annotating data for 3D CAD model and 6D pose, which scales poorly to multiple objects, often rendering it impractical for real applications~\cite{hodavn2019photorealistic, busam2020like, wang2020self6d}. In contrast, our method only needs a short live demonstration of an human-object interaction in order to reliably interact with novel objects.

\subsection{Model-free Grasping}
Model-free approaches are generally very attractive compared to model-based approaches due to their ability to generalize to previously unseen geometry. In essence, model-free grasping can be divided into discriminative approaches~\cite{mousavian20196, mahler2016dex} and generative approaches~\cite{redmon2015real, lenz2015deep}. Discriminative methods sample grasping instructions which are then scored by a neural network, whereas generative models directly output grasping configurations.

Different training data modalities are used to train discriminative approaches. Dex-Net~\cite{mahler2016dex,mahler2017dex} collects a large amounts of samples from a simulator, which is then used to train their proposed Grasp Quality Convolutional Neural Network (CNN). The authors further extend Dex-Net to support suction grippers\cite{mahler2018dex} and dual-arm robots\cite{mahler2019learning}. While the previous methods all require depth data, \cite{levine2018learning} and \cite{pinto2016supersizing} only use RGB inputs to score grasp candidates. To train on real data, Levine \emph{et al.}~\cite{levine2018learning} collect over 800k real grasps over the course of two months. They then train a CNN which scores the grasp success probability given the corresponding motor command. In \cite{satish2019policy} the authors use deep learning to train a robot policy which is capable of fast evaluation of millions of grasp candidates. Mousavian \emph{et al.}~\cite{mousavian20196} leverage a variational auto-encoder to map from partial point cloud observations to a set of possible grasps for the object of interest.
In contrast to our proposed method and generative approaches, discriminative models are usually computationally expensive as each of the proposals needs to be evaluated before grasping can be attempted.

As for generative models, Jiang \emph{et al.}~\cite{jiang2011efficient} infers oriented rectangles in the image plane, representing plausible gripper positioning. Redmon \emph{et al.}~\cite{redmon2015real} simultaneously predict object class and, similar to \cite{jiang2011efficient}, oriented rectangles depicting the grasping instructions. They further extend their method to return a set of multiple possible grasps as most objects can be usually grasped at several locations. Lenz \emph{et al.}~\cite{lenz2015deep} propose a cascade network, where the first network produces candidate grasps, which are subsequently scored by the second network. In \cite{morrison2018closing,morrison2020learning}, the authors leverage a small CNN to generate antipodal grasps. They predict the grasping quality, angle and gripper width for each individual pixel. S$^4$G use a Single-Shot Grasp Proposal Network grounded on PointNet++~\cite{qi2017pointnet} to efficiently predict amodal grasp proposals~\cite{qin2020s4g}. Generative methods are thereby usually tailored towards the robot or gripper used during training. In contrast, we infer the full hand-object interaction. Given the appropriate mapping from human hand to gripper, this can be applied to various robots and grippers.

A few methods also rely on reinforcement learning to teach grasping to a robot~\cite{kalashnikov2018qt, james2019sim, levine2018learning, song2020grasping}. However, reinforcement learning based approaches require a large amount of training data and feedback from real robots to learn grasping of objects. To simplify the training process and enable the robot to grasp specific novel objects, several works utilize human demonstration to learn robotic grasping.  Early work such as \cite{pastor2011online}, \cite{calinon2009learning} store trajectories in robot configuration-space during the demonstration phase. The trajectories are recorded using either kinesthetic teaching \cite{pastor2011online} or teleoperation \cite{calinon2009learning}.  Follow-up papers \cite{yu2018one}, \cite{song2020grasping}
 adopt deep learning approaches to solve the task. Similarly, Yu \emph{et al.}~\cite{yu2018one} teach the robot to perform tasks from a demonstration video. In this work, the robot policy is directly predicted from hidden layers of the network. In comparison, 
 \demograsp\, focuses on inferring the detailed hand pose and object geometry from the network. Although in this paper we transform the human hand pose to 2-finger gripper pose for evaluation purposes, our pipeline has the potential to be leveraged for dexterous robotic hands in humanoid robots.

\section{METHODOLOGY}

\begin{figure*}[t!]
    \centering
    \includegraphics[width=0.975\linewidth]{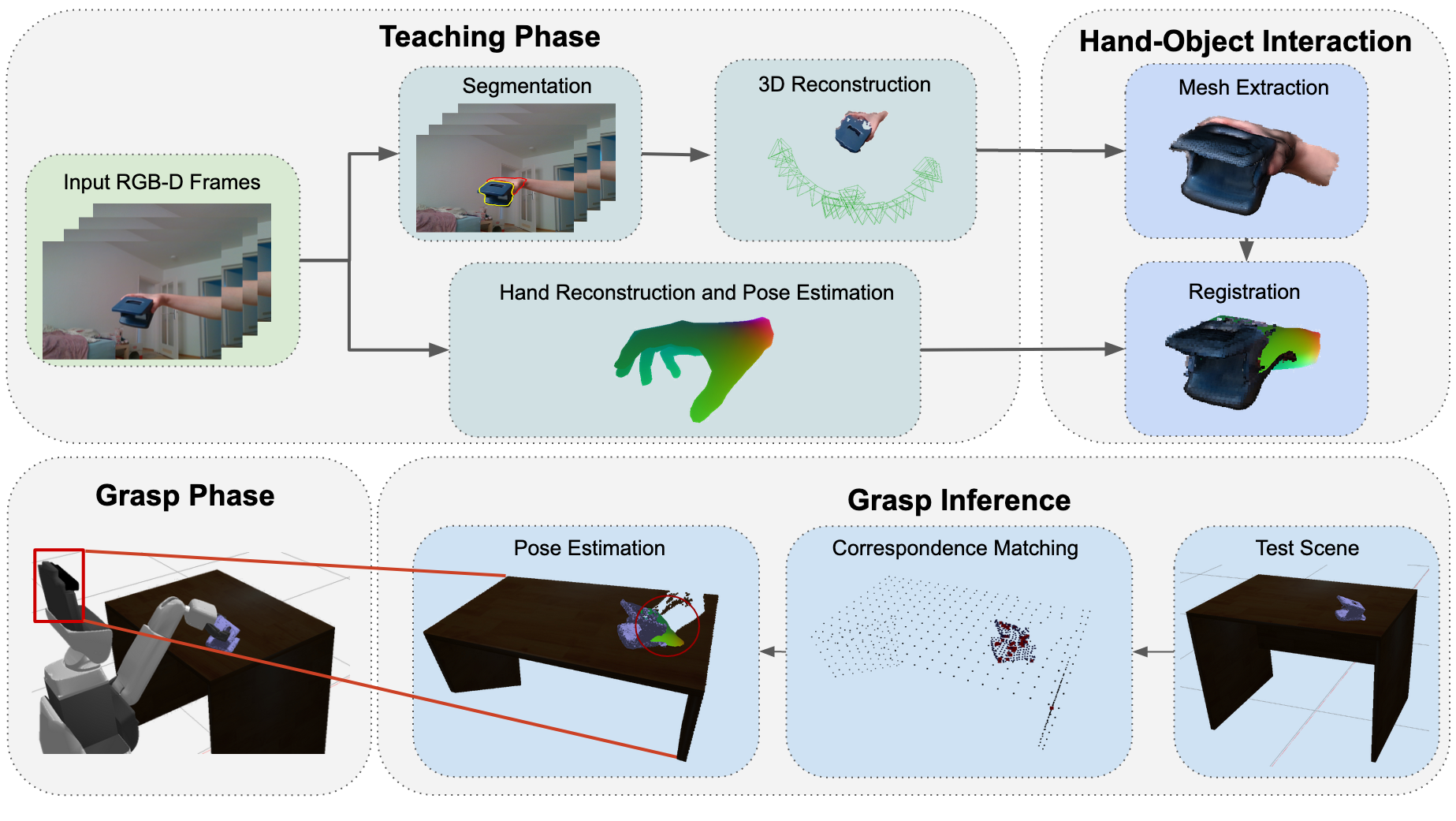}
    \caption{\textbf{Schematic overview of our proposed methodology.} Teaching Phase: After human demonstrations, the hand-object interaction is reconstructed by fusing the RGB-D images pairs  from multiple views with the help of segmentation masks. In the meanwhile, the MANO hand pose ~\cite{hasson2019learning} is retrieved from the first RGB image. After registration, the grasp pose is learned as well as the object shape. Grasp Inference: Points from the object and the scene are matched with PPF-FoldNet ~\cite{deng2018ppf} features, after which the object is registered to the scene. Leveraging the grasp pose registered with the object in the scene, the robot is able to grasp the object with its gripper.
    }
    \label{fig:method}
\end{figure*}
Our \demograsp\, pipeline consists of four consecutive steps, detailed hereafter:
In stage one we segment the hand and object on a set of human demonstration RGB-D images and reconstruct their incomplete shapes leveraging the recorded depth maps.
In stages two and three, we complete the shape of the object and fit a hand pose.
In the last stage, we estimate the 6D pose of the current object in the robot's view which enables transforming the hand model accordingly. The model can subsequently be used to infer plausible grasping instructions for the robot.
See Fig.\ref{fig:method} for a schematic overview of our method.

\subsection{3D Reconstruction of Human-Object Interaction}
To reconstruct the shown hand-object interaction, we first segment the hand and object using Mask R-CNN~\cite{he2017mask}. In the absence of a dataset with real labels for human-object interaction, we rely on synthetic images from the ObMan dataset~\cite{hasson2019learning}.
To enhance segmentation quality, we follow~\cite{lin2017focal} and apply a binary cross entropy for each class independently, thus preventing inter-class competition.
During the demonstration phase we feed our trained Mask R-CNN with the RGB image sequence as perceived by the robot camera.
The segmented hand and object depth images are back-projected into 3D where KinectFusion~\cite{izadi2011kinectfusion} creates a corresponding TSDF volume.
Drift free tracking is achieved by continuously aligning the current input frame against the TSDF volume using ICP~\cite{besl1992method,rusinkiewicz2001efficient}.
Nevertheless, as most household objects often possess rather simple geometry, camera tracking can become challenging. 
We thus simultaneously track the hand and object together in a shared TSDF as the additional hand geometry significantly stabilizes the process and makes the camera pose estimation more robust. 
Finally, we separate hand from object by means of two individual TSDFs reconstructions using the previously estimated relative poses together with the segmentation results.
The additional information from the 2D segmentation is essential in this process as the dynamic motion of the human demonstrator's arm and body in relation to the static environment can be filtered out as to solely fuse the relevant hand and object information.

\subsection{Object Shape Completion}
After reconstruction, the object model is not yet complete due to self-occlusion and partial visibility. 
As a detailed and complete mesh has a positive influence on the subsequent tasks, we first apply shape completion.
While there are different approaches to achieve this, in this work we harness a 3D CNN to directly correct the TSDF volume, prior to shape extraction via marching cubes~\cite{lorensen1987marching}. 
Despite the cubic complexity scaling of TSDFs, they still provide a suitable solution in our case where both object and hand have a limited spatial extent.
We apply a 3D variant of U-Net~\cite{ronneberger2015u} with skip-connections for improved accuracy and feed it with the fused object TSDF volume of $64\times64\times64$ resolution. 
The 3D volume is encoded to a 512-dimensional feature descriptor which is additionally concatenated with RGB features to provide complementary image information in addition to the geometry.
The decoder consists of 6 up-convolutions with a kernel size of $3\times3\times3$ followed by a max-pooling layer to eventually reach the input resolution.
The object mesh prediction is modeled as a classification task where each voxel in the output volume represents its prediction score, denoting if a voxel is occupied.
Since only a small portion of voxels in the volume is occupied, the training data is highly imbalanced. We thus apply a class average loss and additionally use the focal loss as objective function to re-balance the loss contributions~\cite{lin2017focal} according to
\begin{equation}
\begin{split}
    L(p) = & \frac{1}{\text{Pos}}\sum_{p \in \text{Pos}}-(1-p)^\gamma \log (p)\\ & + \frac{1}{\text{Neg}}\sum_{p \in \text{Neg}}-(1-p)^\gamma \log (p).
\end{split}
\end{equation}
where $\text{Pos}$ and $\text{Neg}$ respectively represent the occupied and empty voxels and $\gamma = 2$ as originally proposed by Lin \emph{et al.}~\cite{lin2017focal}.
To train our shape completion model we render multiple images from different views of various hand-object interactions using the synthetic simulator of~\cite{hasson2019learning}. We then extract the associated object TSDFs using the previously described procedure. These extracted object TSDFs are eventually fed into the network as training input, while the object meshes from simulation are transformed to their voxel representation in order to obtain the associated groundtruth.

\subsection{Hand Pose Estimation}
We retrieve parameters for a statistical hand model from the reconstructed hand shape.
To this end we use the MANO~\cite{romero2017embodied} hand model which maps parameters for hand pose and shape to a mesh. The MANO hand model is developed from multiple scanned real hands featuring realistic deformation.

\begin{figure}[t!]
 \centering
    \includegraphics[width=0.4\linewidth]{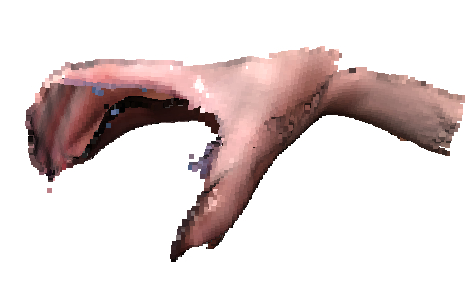}
    \includegraphics[width=0.4\linewidth]{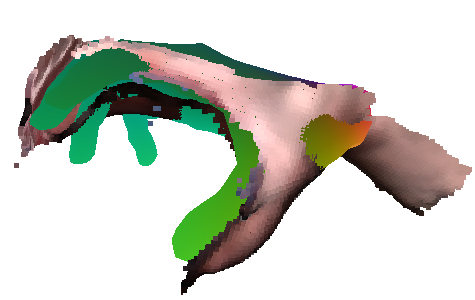}
    \caption{\textbf{Registration of MANO Hand.} Left: Fused hand point cloud. Right: The MANO hand mesh (in green) registered to the point cloud.}
    \label{hand_icp}
\end{figure}

As prediction under occlusion is difficult, the authors of \cite{hasson2019learning} propose to jointly train their CNN for hand pose and object mesh estimation using auxiliary contact and collision losses to encourage natural predictions without collision, which mutually improves both tasks.
We eventually leverage the hand branch of their architecture for a complete hand mesh reconstruction. As the hand is part of a joint reconstruction, the pose is already close to the actual grasping location. Nevertheless, to further improve grasping, we additionally employ ICP to tightly align the hand mesh with the point cloud extracted from the TSDF volume of the partial hand reconstruction.
This allows to calculate the transformation of the hand mesh in robot view coordinates ready to calculate grasping instructions.
The final hand and object mesh after alignment is visualized in Fig.~\ref{hand_icp}.

\subsection{Grasping Point \& Instruction Retrieval}
After accurate calculation of both object and hand shape from the demonstration, we want to determine the grasping instructions given the robot view.
We first retrieve the object pose which is then used to transform the hand mesh and calculate the grasping points with thumb and index finger of the hand model.
The first step is realized with PPF-FoldNet~\cite{deng2018ppf} which encodes the information of the local geometry in a high dimensional feature descriptor.
Thereby, we calculate features for both the demo reconstruction and the current robotic observation.
We then align both point clouds through robust feature matching using RANSAC.

As our extracted hand mesh is aligned with the object pose, we can infer the associated human grasp for the given input.
Now that we estimated hand and object pose in the robotic view coordinates, we can determine the grasping instructions depending on the gripper hardware.
We exemplify this with the Toyota Human Support Robot (HSR) and a two-sided gripper in the evaluation section below. 

\section{EVALUATION}
We evaluate \demograsp\, in both simulation and the real world. To enable a comparison with state-of-the-art, we introduce a new synthetic dataset and an evaluation benchmark which we make publicly available.

\subsection{Evaluation Protocol}

\begin{figure}[t!]
    \centering
    \includegraphics[width=0.24\linewidth]{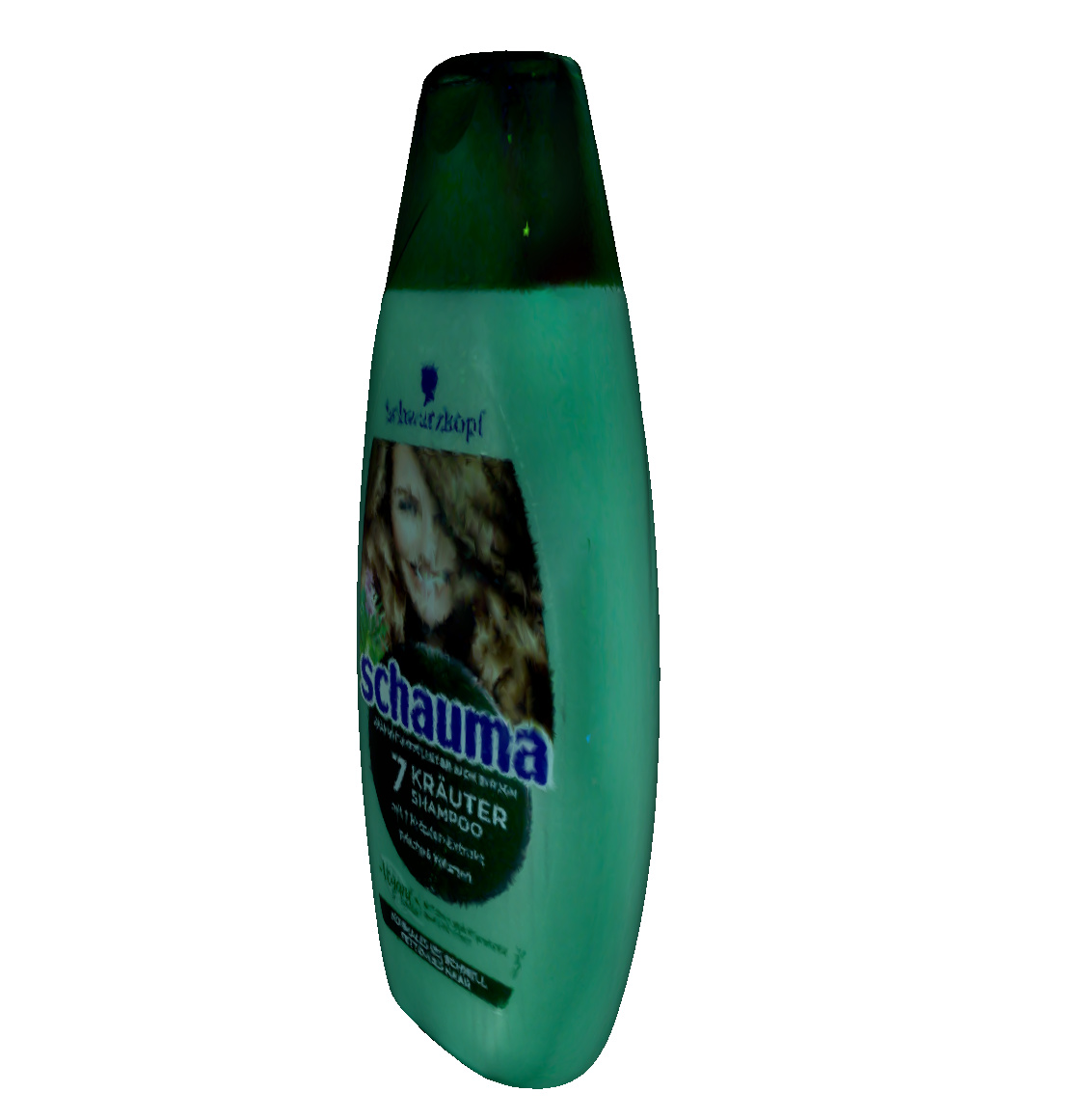}
    \includegraphics[width=0.24\linewidth]{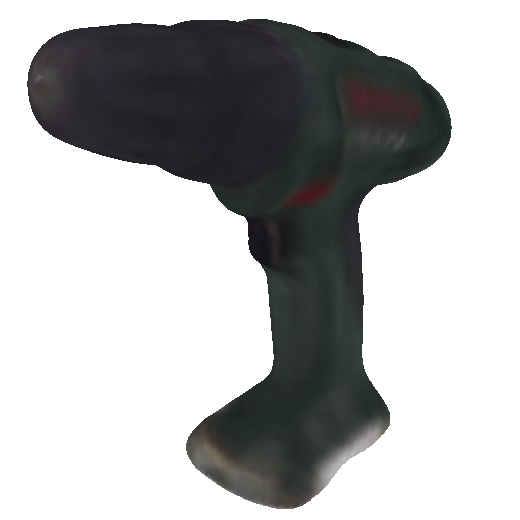}
    \includegraphics[width=0.24\linewidth]{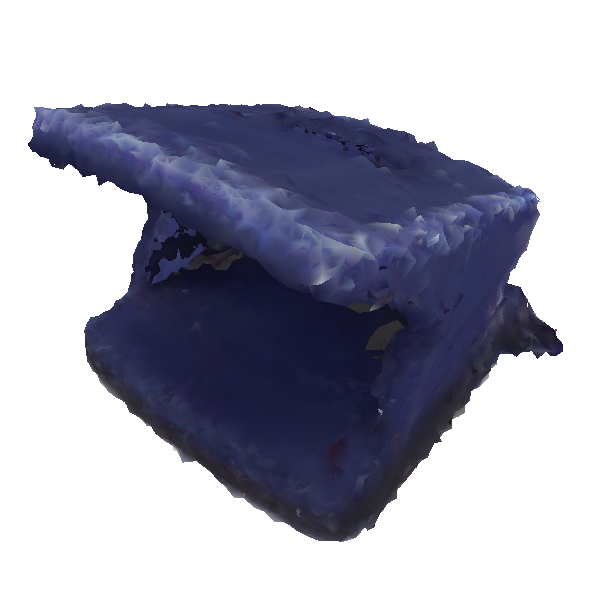}
    \includegraphics[width=0.24\linewidth]{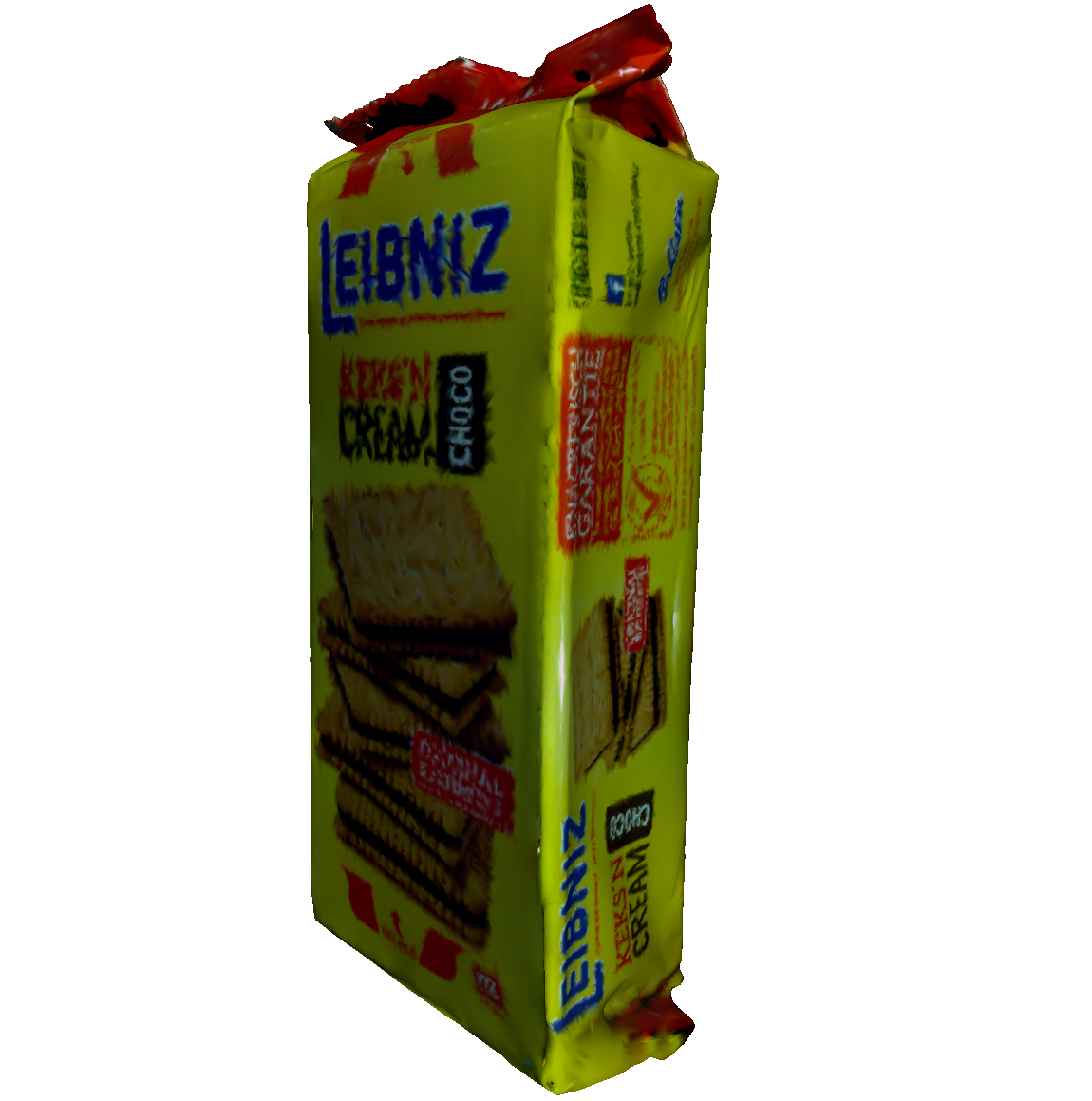}
   \caption{\textbf{Test objects used in the simulation.} From left to right: shampoo bottle, drill, hole punch, cookie box.}
    \label{fig:synthetic_test_objects}
\end{figure}

We select four objects of different shape --- drill, hole punch, cookie box and shampoo bottle --- for our in silico tests (see Fig.~\ref{fig:synthetic_test_objects}). To obtain the ground truth object shapes, we scanned the cookie box and shampoo bottle using a 3D scanner (Shining3D EinScan-SP), and acquired the hole punch and drill mesh model from LineMOD dataset~\cite{hinterstoisser2011multimodal}. For evaluation we use Toyota HSR, a service robot platform that provides several integrated functionalities for a wide range of applications. We leverage the open-sourced simulation environment for HSR on which we build our benchmark \cite{yamamoto2019development}.

\begin{figure*}[t!]
    \centering
    \includegraphics[width=0.19\linewidth]{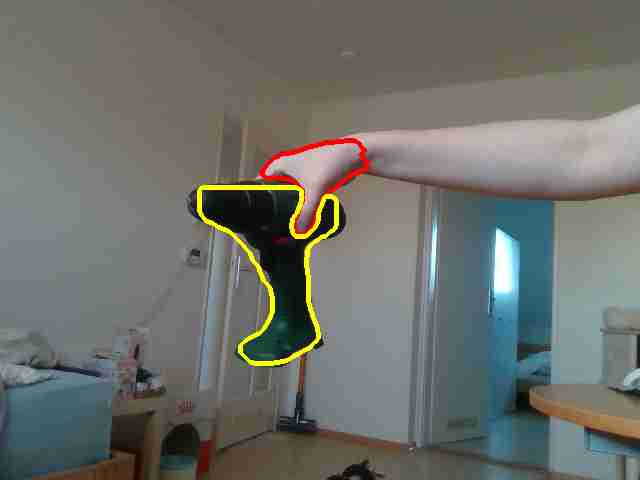}
    \includegraphics[width=0.19\linewidth]{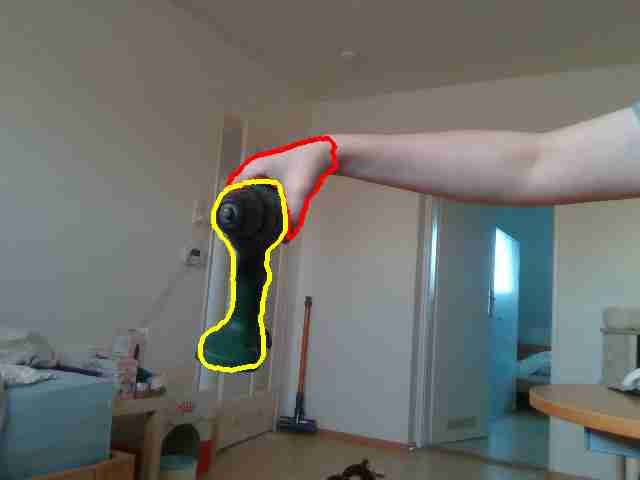}
    \includegraphics[width=0.19\linewidth]{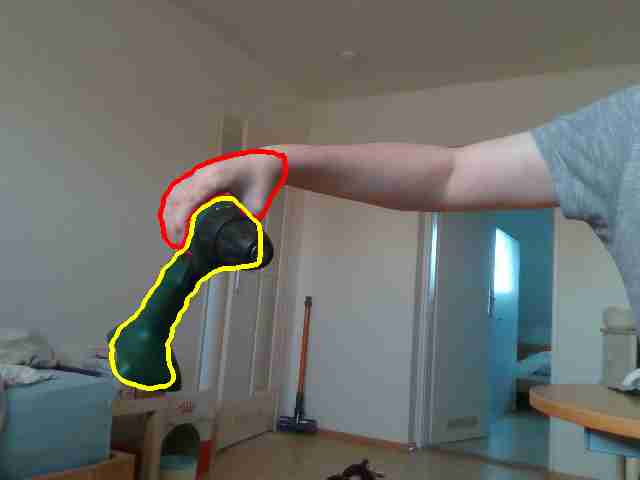}
    \includegraphics[width=0.19\linewidth]{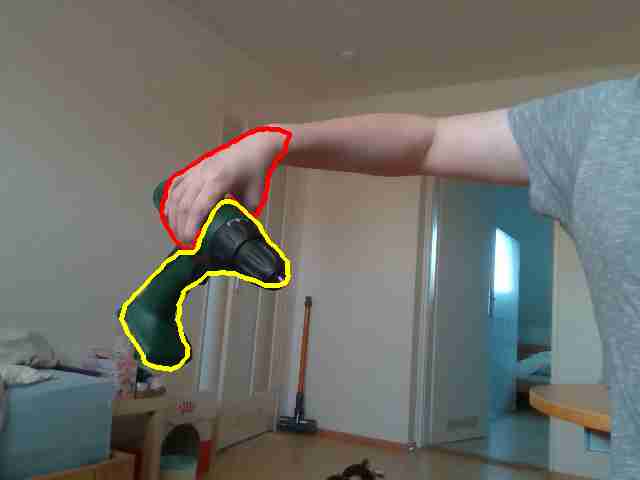}
     \vline 
     \includegraphics[height=2.4cm]{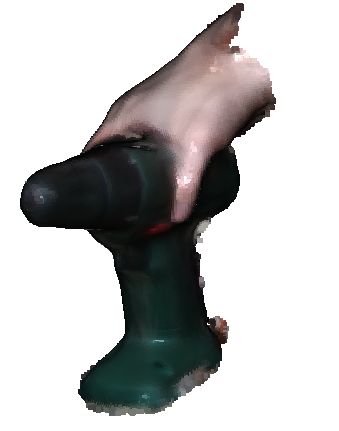}
   \caption{\textbf{Exemplary training images.} We show four example images from the demonstration sequence of the dill object together with our segmentation results for hand and object, which are used to reconstruct the overall interaction (right). The demonstration sequences are in average 10-30 seconds long and cover around 100-400 images.} %
    \label{fig:training_sequence}
\end{figure*}

Each test scene is set up with a table one meter away in front of the robot, in the view of the robot head RGB-D camera. The scanned object models are placed at different locations and viewing angles on the table. We consider a grasp successful only if the robot is able to grasp the object with its gripper, retrieve it from the table, and hold it still for three seconds without dropping it. In addition, we also record a human demonstration for each object, which we release with the benchmark, consisting of short RGB-D sequences of less than 1 minute length ($\approx$ 100-400 frames) demonstrating a hand-object interaction at a distance of 0.5~m. During demonstration we rotated the objects once around by hand as shown in Fig.~\ref{fig:training_sequence}.

\begin{figure}[t!]
  \centering
  \includegraphics[width=0.8\linewidth]{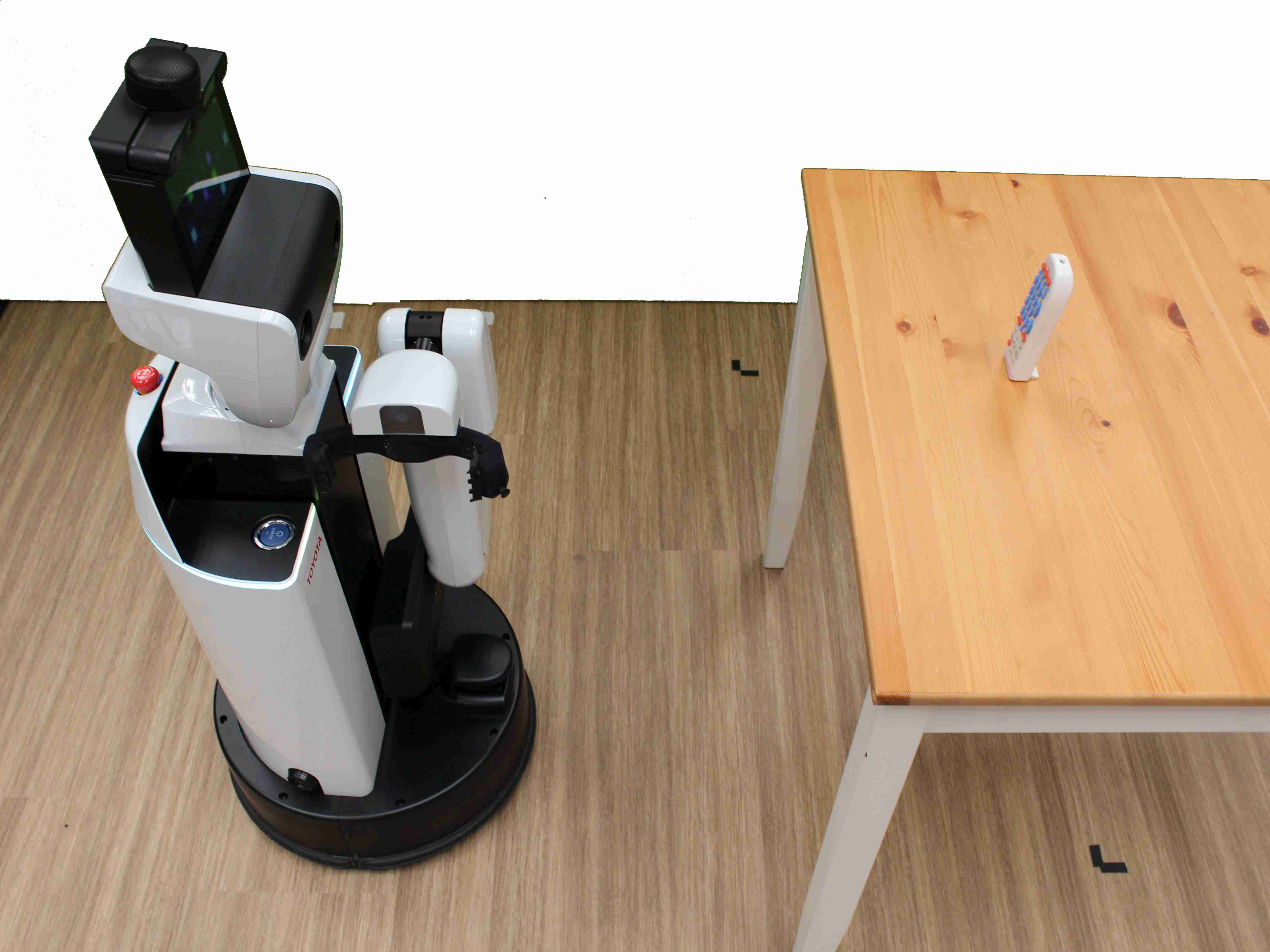}
\caption{\textbf{Our experimental setup.} We randomly place a demonstrated object on the table in front of HSR. Afterwards, we let HSR retrieve the object using the grasp instructions inferred by our \demograsp\, pipeline.}
\label{fig:exp_setup}
\end{figure}

As for the real experiments, we evaluate our pipeline in a physical grasping task using Toyota HSR. The goal of the task is again for HSR to grasp a previously-learned household object and retrieve it from the table. We use four household objects: a cup, a can, a bottle, and a remote control (see Fig.~\ref{fig:results_hsr}). For each of these objects, we record two human demonstration sequences, one in which the demonstrator holds the object from its side, and one from the top. For each demonstration sequence, the experiments are conducted as follows: we record the RGB-D demonstration sequence using HSR's head camera. Then, we run our pipeline to create a model for the object and the corresponding hand model for grasping the object. Finally, we place the object on a table in front of the robot and let HSR grasp the object and retrieve it from the table. To assess the robustness of the learned models, we vary the placement of the objects on the table. In particular, we divide the front half of the table --- that is, the closer half of the table w.r.t. the initial position of HSR -- in three equal areas. In each run, we select one of the three areas, randomly place the object there and let HSR attempt to grasp it. We repeat the grasping experiment three times for each of the areas for each demonstration sequence. We thus perform 9 runs per demonstration sequence, 18 runs per object, and 72 runs overall. A run is considered successful if HSR is able to grasp the object, retrieve it from the table, and move back to a neutral position with the object still in its gripper. Notice that the rear half of the table is excluded from the experiments, as reaching it would create challenges for the path planner of HSR. A representation of the experimental environment is provided in Fig.~\ref{fig:exp_setup}.

\subsection{Extracting Grasping Instructions for Toyota HSR}
Given the hand predicted in the test scene by \demograsp, we generate the target 6D grasp pose for HSR's gripper based on the locations of the index finger, the thumb, and the wrist. The final grasp location is then calculated as the 3D middle point between the index and thumb locations. 

\begin{figure}[t!]
\centering
\includegraphics[height=3.3cm]{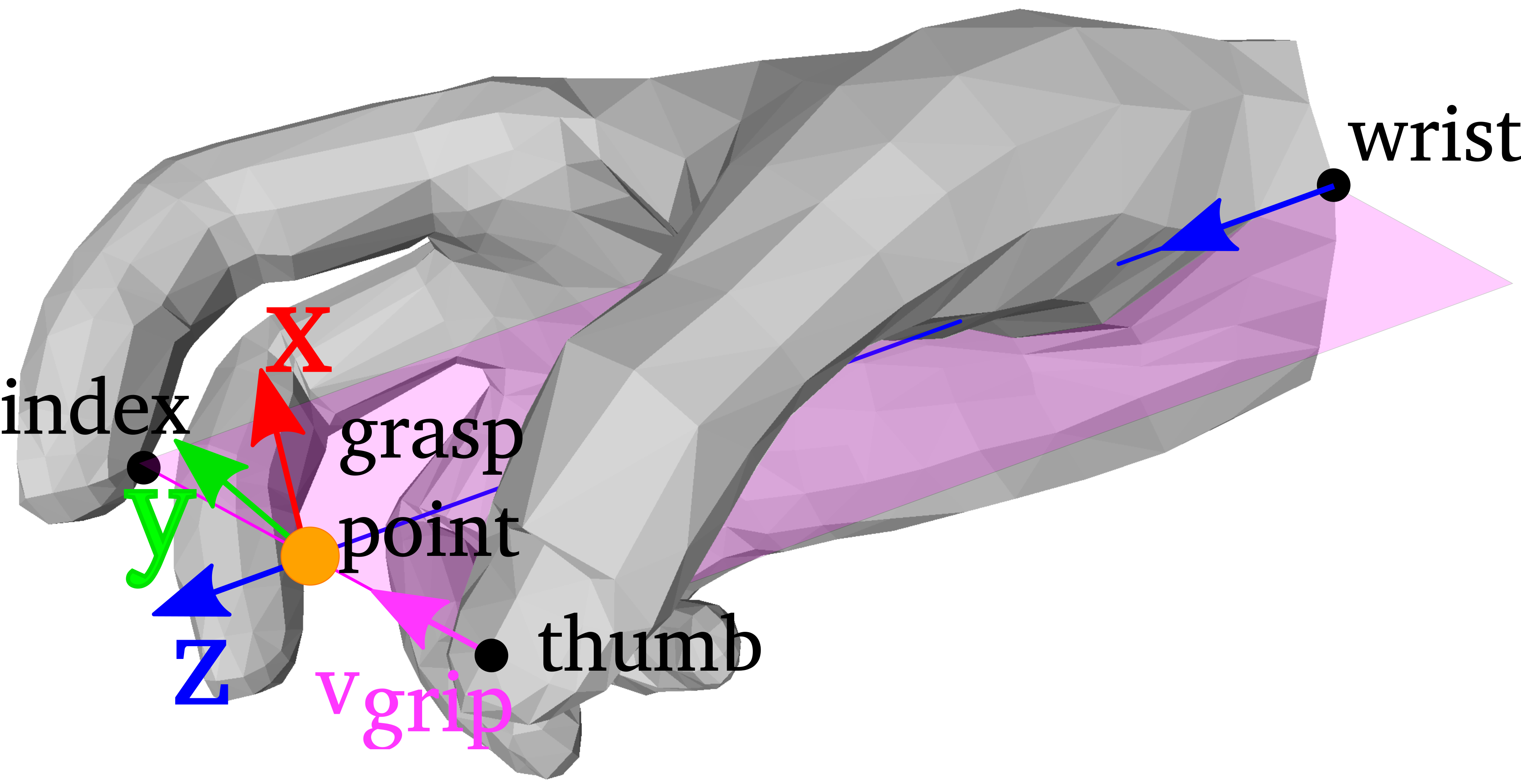}
\includegraphics[height=2.75cm]{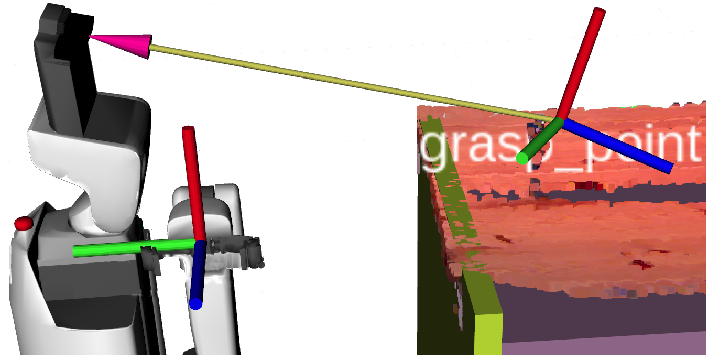}
\caption{\textbf{Calculation of the 6D grasp pose.} Illustration of the logic that we use to calculate the grasp point and rotation given the hand predicted by \demograsp\, (top). The lower part shows a representation of the predicted grasp pose by \demograsp\, in the synthetic environment.}
\label{fig:hsr_vect_logic}
\end{figure}

For rotation, we create a coordinate system using three vectors:
\begin{itemize}
\item The vector $\vec{z}$ between the wrist location and the calculated middle point between the index and thumb locations, which is the vector that HSR's hand will follow to approach the object;
\item The vector $\vec{x}$ perpendicular to the plane described by $\vec{z}$ and the the vector $\vec{v_{grip}}$ between the thumb and the index finger location, which is the direction in which HSR will tighten the grip;
 and
\item The vector $\vec{y}$ perpendicular to the plane described by vectors $\vec{x}$ and $\vec{z}$ as $\vec{y} = \vec{z} \times \vec{x}$
\end{itemize}
An illustration of this logic is provided in Fig.~\ref{fig:hsr_vect_logic}.

\begin{table*}[t!]
    \centering
    \begin{tabular}{llllll}
    \toprule
        Successful Grasps / Average Planning Attempts for & Shampoo & Drill & Hole Punch & Cookie Box & Mean\\
        \midrule
        \graspn~\cite{mousavian20196} & \textbf{86.7} / 5.6  & 73.3 / 7.2 & \textbf{86.7} / 11.2 & 60.0 / 108.3 & 76.7 / 41.8 \\
        \midrule
        \demograsp\,w/o Shape Completion &  73.3 / -- & \textbf{100.0} / -- & 66.7 / -- & \textbf{100.0} / -- & 85.0 -- \\ 
        \demograsp\, &  \textbf{86.7} / -- & 93.3 / -- & \textbf{86.7} / -- & \textbf{100.0} / -- & \textbf{91.7} / --\\
        \bottomrule
    \end{tabular}
    \caption{\textbf{Grasping results on synthetic evaluation suite.} We compare \demograsp\, with the state-of-the-art method \graspn~\cite{mousavian20196}. We additionally report the average number of attempts for the motion planning in \graspn, as \cite{mousavian20196} infers several possible grasps which are consecutively ranked. The second row shows an ablation of our \demograsp\, pipeline without the shape completion module.}
    \label{tab:synthetic_results}
\end{table*}

\subsection{Evaluation Results in Simulation}
During evaluation we use the aforementioned training sequences to train our method. We then run \demograsp\, according to the described evaluation protocol and report the average success rate after 15 different grasping trials in TABLE~\ref{tab:synthetic_results}. In particular, we note robust grasping results with an average success rate of $91.7\%$ over all objects, clearly showing that \demograsp\, is capable of producing reliable grasps. Disabling the shape completion module, we saw a decrease in our success rate of approximately $6\%$, which indicates the importance of completion for reliable grasping instructions. Interestingly, we are on par or better for all objects when using shape completion except for the drill. As this object is fairly large, matching results with PPF-FoldNet are reliable even without completion. Completion can add some noise in this case which lowers the performance for the large object. Nevertheless, for most hand-sized objects such as shampoo and hole punch, our results with shape completion are clearly superior, as the hand covers most of the object and thus matching without the completion cannot be reliably conducted.

We additionally compare \demograsp\, with \graspn\,, a state-of-the-art approach for model-free grasping. We outperform \graspn\, by $15.0\%$ with an average success rate of $76.7\%$ compared to our $91.7\%$. Moreover, \graspn\, is a discriminative approach that samples up to 900 plausible grasps with an associated confidence. These grasp poses are then sorted from high to low scores and tested with a motion planning algorithm one after the other to assess if the pose is feasible. In our experiments, \graspn\, tested 41.8 candidate grasps per trial, on average.
In contrast, our method evaluates only one single grasp pose, which makes the motion planning on average about 40 times faster.
To enable future comparison, we will publicly release our benchmark suite as well as the training sequences that we recorded.

\subsection{Real World Evaluation Results on Toyota HSR}

\begin{figure}[t!]
\centering
\begin{tabular}{ccc}
    \toprule
     \multicolumn{3}{c}{Success Rate} \\
     \midrule
     grasp from side & grasp from top & mean \\
     \midrule
     94.4 & 58.3 & 76.4\\
     \bottomrule
\vspace*{2mm}
\end{tabular}
\includegraphics[width=0.99\linewidth]{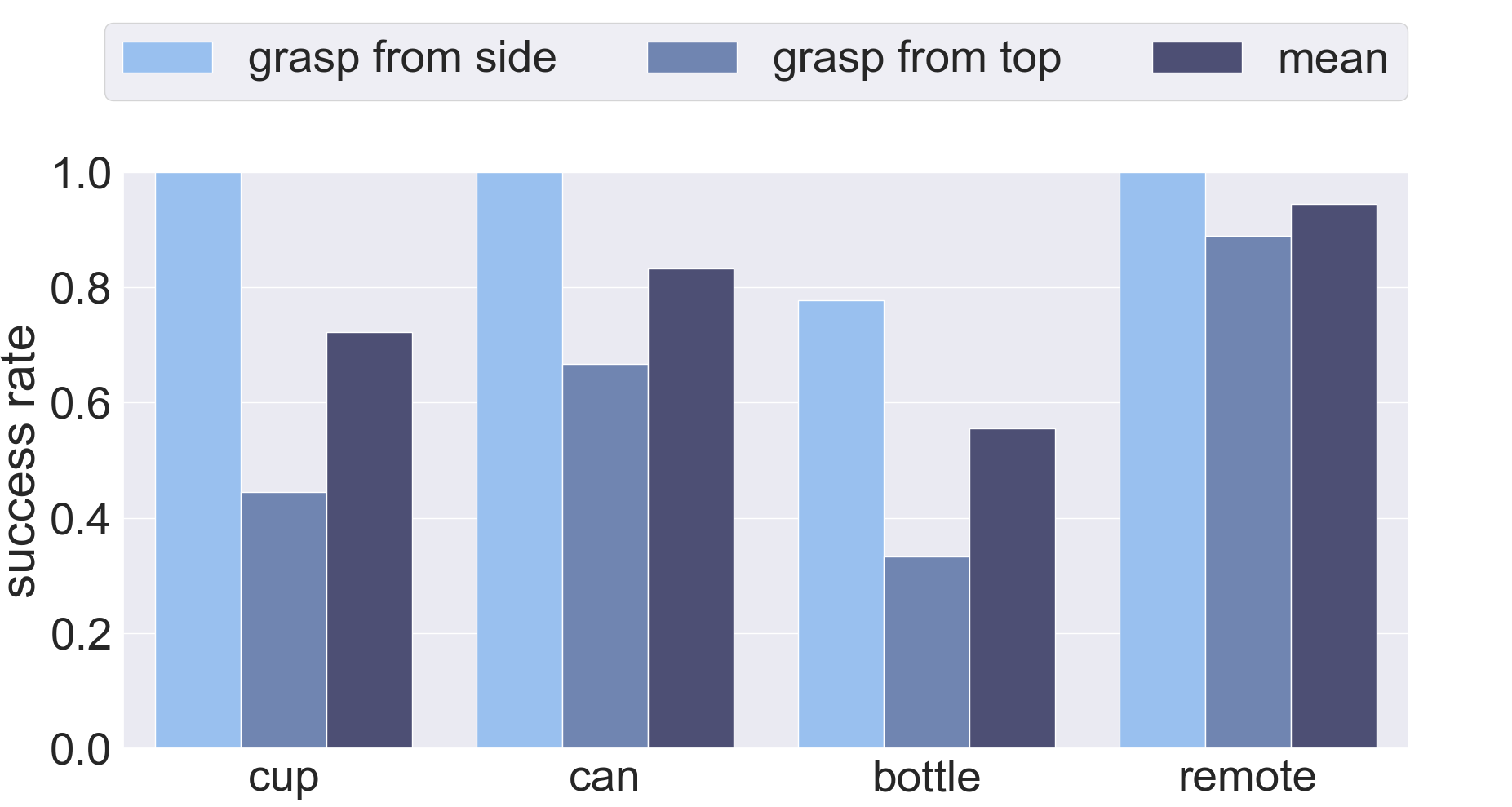}
\includegraphics[width=0.83\linewidth]{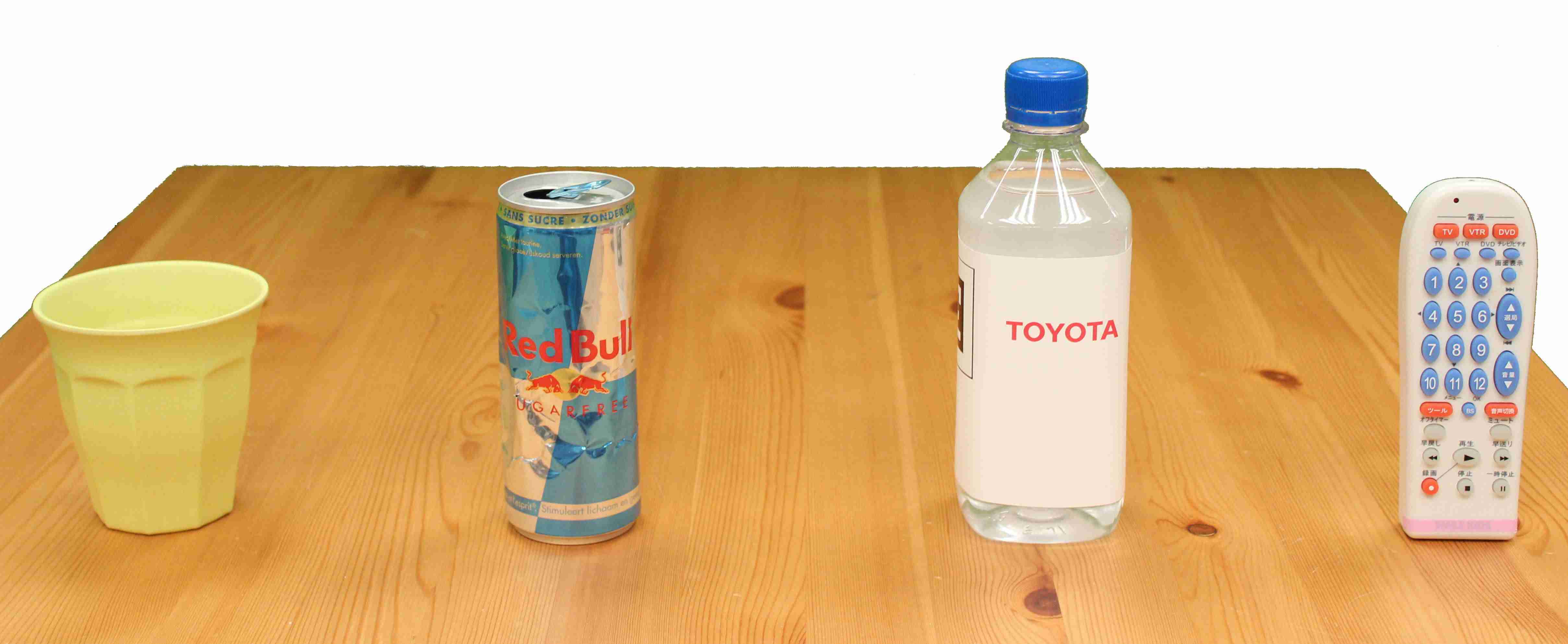}
\caption{\textbf{DemoGrasp success rate on a real HSR.} Overall success rate for different grasp sides (top). The diagram shows individual grasping results for each object when they are grasped from the side and from top (middle). The four objects that we used for our real-world experiments (bottom).}
\label{fig:results_hsr}
\end{figure}

Fig.~\ref{fig:results_hsr} summarizes the results of our real experiments. The grasp success rate is $94.4\%$ when grasping objects from their side, while a grasp from the top is successful $58.3\%$ of the time. We postulate that this difference is due to the fact that 
\begin{inparaenum}[i)]
\item most of the objects considered are horizontally symmetrical, which sometimes can cause our pipeline to place the predicted hand at the bottom rather than at the top of the object, eventually leading to a grasping failure, and
\item the considered objects offer a larger surface for grasping on their side compared to their top.
\end{inparaenum}
The drop in success rate for the bottle is mostly due to the partial transparency of the object, which makes the object shape reconstruction, and consequently the registration, more challenging. 
Nonetheless, HSR was able to successfully grasp the object in $76.4\%$ of the runs simply by observing a single interaction between a person and the object. In addition, a live demo of the real experiment is shown in the attached video.

\begin{table}[t!]
    \centering
    \begin{tabular}{llllll}
    \toprule
        \multicolumn{5}{c}{Success Rate} \\
        \midrule
        Demo & \multirow{2}{*}{Shampoo} & \multirow{2}{*}{Drill} & \multirow{2}{*}{Hole Punch} & \multirow{2}{*}{Cookie} & \multirow{2}{*}{Average} \\
        Sequence & & & & \\
        \midrule
        100\% &  86.7  & 93.3  & 86.7 & 100.0 & 91.7\\
        \midrule
        50\%  &  80.0  & 100.0  & 60.0  & 33.3 & 68.3\\ 
        \midrule
        30\%  & 66.7   & 100.0 & 13.3 & 6.7 & 46.7\\
        \bottomrule
    \end{tabular}
    \caption{\textbf{Grasping results for learning sequences with partial visibility.} We reduce the demonstration sequence length and compare the corresponding \demograsp\, success rates in our simulation environment.
    }
    \label{tab:fraction_results}
\end{table}

\subsection{Restricting the Learning Sequence to partial Visibility}
During the teaching phase, \demograsp\,  records and fuses images by demonstrating the object in a single motion from one side to the other providing different views of the hand-object interaction.
To understand the effect and limitation of partial hand-object demonstration on the result of \demograsp, we run the teaching phase for our simulated grasp experiments with decreasing amount of demonstration length using different fractions of the original demonstration sequences.
For each of the fractions, we then run the grasp experiments with the Toyota HSR in our simulation environment, and compare the success rates in TABLE~\ref{tab:fraction_results}. It is worth to note that random seeding for RANSAC caused one grasping mistake in our test for the large drill object.
The results show that using fewer frames as learning sequence leads to a drop in the average success rate for the objects by $23.4$ points using half of the demonstration video and $45.0$ points with only $30\%$.
The incomplete demonstration of both the object and the hand geometry results in a less accurate reconstruction, which has a direct effect on the success of the entire pipeline eventually leading to a failure in registration of the reconstructed MANO hand model with the hand point cloud which severely affects the grasping of the objects hole punch and cookie box.
We observe that demonstration sequences that show all the fingers of the hand at least once to the camera are key to robust grasping. %

\section{CONCLUSIONS}
In this paper we introduced \demograsp, a novel method for inferring grasping instructions from a short human demonstration. In the core, a human demonstrates a hand-object interaction in a short RGB-D image sequence. The sequence is leveraged to jointly reconstruct the 3D object as well as the hand mesh. Then, we localize the object in 3D using point-pair features and estimate reliable grasping instruction from the previously reconstructed interaction. Exhaustive evaluations in a real and synthetic environments demonstrate the applicability of our approach to learning grasping instructions from a single human demonstration.
In the current pipeline, our model is restricted to the prediction of a single object in the scene.
Future research will focus on extending \demograsp\, with incremental learning: the grasping instructions for new objects can be incrementally learned and added to the existing object library.
Overall, we believe that our methodology can pave the way towards natural and interpretable human-robot collaboration by imitation, and our new dataset can serve as a basis to compare future approaches.

{\small
\bibliographystyle{IEEEtran}
\bibliography{egbib}
}

\end{document}